\documentclass{article}

\usepackage[final]{neurips_2018}

\usepackage[utf8]{inputenc} 
\usepackage[T1]{fontenc}    
\usepackage{hyperref}       
\usepackage{url}            
\usepackage{booktabs}       
\usepackage{amsfonts}       
\usepackage{nicefrac}       
\usepackage{microtype}      

\usepackage{graphicx}
\usepackage{caption}
\usepackage{subcaption}
\usepackage{amsmath} 
\usepackage{float} 
\usepackage{bm}
\usepackage{tikz}

\usepackage{pgfplots}
\usepgfplotslibrary{groupplots}
\usepgfplotslibrary{dateplot}
\usepackage{sansmath}
\pgfplotsset{
  compat=newest,
  tick label style = {font=\sansmath\tiny},
  every axis label = {font=\sansmath\small},
  legend style = {font=\sansmath\tiny, fill opacity=0.8},
  label style = {font=\sansmath\small}
}

\usepackage{tikzsymbols} 

\setcitestyle{authoryear,open={(},close={)}}
\title{Loss Patterns of Neural Networks}

\author{
  Ivan Skorokhodov \\
  Neural Networks and Deep Learning Lab \\
  Moscow Institute of Physics and Technology \\
  \texttt{skorokhodov.is@mipt.ru} \\
  \And
  Mikhail Burtsev \\
  Neural Networks and Deep Learning Lab \\
  Moscow Institute of Physics and Technology \\
  \texttt{burtcev.ms@mipt.ru} \\
}

\begin{document}

\maketitle

\begin{abstract}
We present \textit{multi-point optimization}: an optimization technique that allows to train several models simultaneously without the need to keep the parameters of each one individually. The proposed method is used for a thorough empirical analysis of the loss landscape of neural networks.
By extensive experiments on FashionMNIST and CIFAR10 datasets we demonstrate two things: 1) loss surface is surprisingly diverse and intricate in terms of landscape patterns it contains, and 2) adding batch normalization makes it more smooth.
Source code to reproduce all the reported results is available on GitHub\footnote{\url{https://github.com/universome/loss-patterns}}.
\end{abstract}

\section{Introduction}

\begin{figure}[H]
    \centering
    
    
    \begin{subfigure}{0.48\textwidth}
        \centering
        \includegraphics[height=2.5cm]{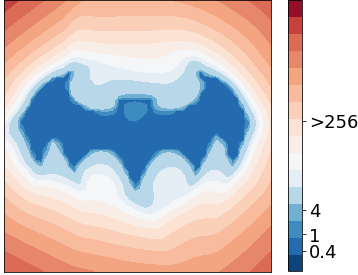}
        \includegraphics[height=2.5cm]{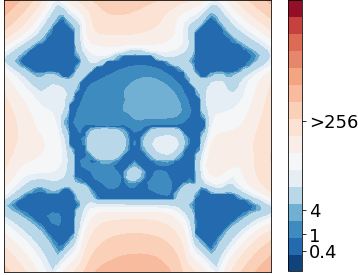}
        \caption{Loss surface on FashionMNIST dataset}
    \end{subfigure}
    \begin{subfigure}{0.48\textwidth}
        \centering
        \includegraphics[height=2.5cm]{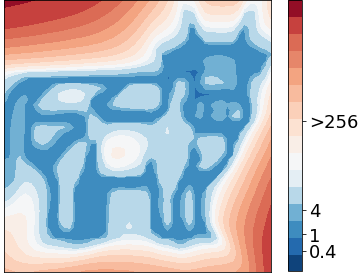}
        \includegraphics[height=2.5cm]{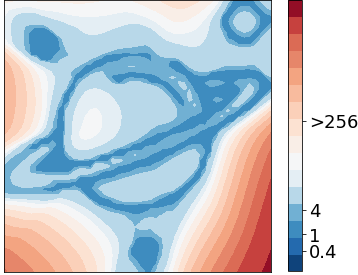}
        \caption{Loss surface on CIFAR10 dataset}
    \end{subfigure}
    \caption{Examples of a loss landscape of a typical CNN model on FashionMNIST and CIFAR10 datasets found with MPO. Loss values are color-coded according to a logarithmic scale. We used a small VGG-like model which architecture is presented in appendix \ref{appendix:hyperparams}; additional visualizations are presented in appendix \ref{appendix:visualizations}. Here we used \textit{test} sets to compute the loss values.}
    \label{fig:icons-grid}
\end{figure}

In this paper, we present \textit{multi-point optimization} (MPO): a technique that allows to find many weight vectors in the parameter space by performing optimization procedure on a considerably smaller amount of parameters, thus saving a lot of memory and computation.
This technique allowed us to explore the structure of a loss landscape of neural networks on FashionMNIST and CIFAR10 datasets by finding different landscape patterns (see, for example, Figure \ref{fig:icons-grid}), and to analyze smoothing capabilities of batch normalization (\citet{BatchNorm}).

\citet{Sublevel_sets} recently showed that, under some mild assumptions, any two minima of a neural network loss surface are connected by a continuous path of approximately the same loss value.
\citet{Mode_connectivity} proposed a practical approach to find such a path.
In our work, we extend this approach and instead of finding parameters of a 1D-path between 2 fixed minima we find parameters of a $d$-dimensional manifold under some generous constraints over $K$ non-fixed weight vectors.

\section{Related work}

Proposed method originates from mode connectivity ideas developed concurrently by \citet{Mode_connectivity} and \citet{Mode_connectivity_via_difficult_way}.
In those papers, authors empirically demonstrate that any two local minima can be connected by a continuous path of approximately the same loss value. \citet{Sublevel_sets} went further and rigorously showed that under very mild assumptions all sublevel sets of loss surface of neural networks are connected. Our approach is based on the optimization procedure used by \citet{Mode_connectivity} to connect the modes.
But instead of finding a single 1D-path between existing 2 weight vectors we fit parameters of a $d$-dimensional manifold under arbitrary constraint over $K$ weight vectors.

To the best of our knowledge multi-point optimization of neural networks was not previously explored in such a general scenario.
The one specific direction in which similar ideas were developed is fast ensembling strategies.
For example, \citet{SnapShot_ensembles, Mode_connectivity} recently showed that one can construct a good ensemble by taking models along the SGD trajectory if one carefully perturbs the learning rate during the optimization process to force the exploration of other minima.

Current work does not address the ensembling potential of the proposed method, and we focus on the loss landscape analysis instead.
In this sense it is highly related to the work of \citet{Visualizing_Loss_Landscape} who performed an empirical investigation of the loss surfaces of large-scale neural networks.
It was demonstrated that deep models tend to have a more irregular loss surface and one of the techniques to make it more smooth is to use skip-connections.
Some analysis of the loss landscape with mode connectivity ideas was performed by \citet{Using_Mode_Connectivity_for_Loss_Landscape_Analysis} who demonstrated that while plain SGD simply goes down the hill --- SGD with warm restarts walks around the barriers.
The fact that batch normalization (BN) smoothes the loss landscape is not new and was previously shown by \citet{How_BN_helps_optimization}.
Our work provides one more empirical validation of this fact by using another set of tools.

\section{Our method}
\label{section:method}
Let the parameter space of the model be $\mathbb{R}^n$.
Multi-point optimization works by arranging $K$ weight vectors $\bm w_1, ..., \bm w_K \in \mathbb{R}^n$ on some $d$-dimensional manifold $M_{\bm{\theta}} \subset \mathbb{R}^n$ and then optimizing its parameters $\bm{\theta} \in \Theta$.
To avoid unnecessary abstraction and complications, and since in all the presented experiments we set $d$-dimensional manifold to be just a 2D-plane, hereinafter we describe our method for $M_{\bm \theta} = \mathbb{R}^2 \subset \mathbb{R}^n$.
But we emphasize that it is not limited to this setup and can be straightforwardly extended to other structures and dimensionalities.

Optimizing parameters of $K$ weight vectors lying on some 2D-plane is not very exciting and one would like to regularize this optimization in some interesting way.
In all our experiments we use a very specific family of constraints over $\bm w_i, i = 1, ..., K$: we force these weight vectors to form some binary 2D-picture.
To save space we allow ourselves to describe our method specifically for this constraint family.
But we again emphasize that other regularization families can be used.
For example, following \citet{Ensembling_decorrelation}, one can force weight vectors to provide models with decorrelated predictions, thus making them form a stronger ensemble.

\begin{figure}
     \centering
     \begin{subfigure}[b]{0.32\textwidth}
         \centering
         \includegraphics[height=2.8cm]{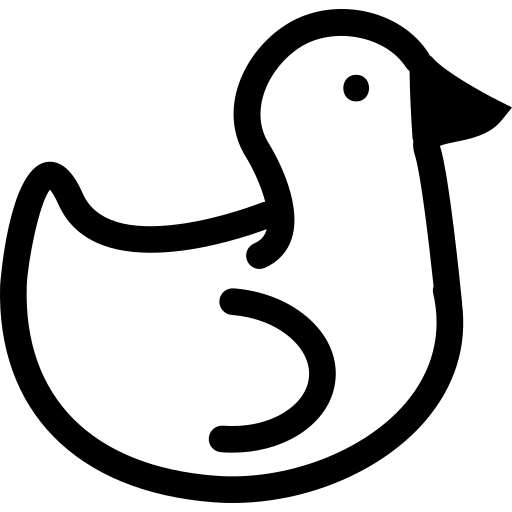}
         \vspace{0.5cm}
         \caption{Original pattern we want to find in the weight space}
         \label{fig:method:duck-icon}
     \end{subfigure}
     \hfill
     \begin{subfigure}[b]{0.32\textwidth}
        \centering
        \begin{tikzpicture}
        \tikzset{>=latex}
        \tikzstyle{every node}=[font=\small]
        \node[font=\tiny] at (1.5,0.4) {$w_{28,13} = w_O + 28 \cdot w_\text{up} + 13 \cdot w_\text{right}$};
        \node[inner sep=0pt] (whitehead) at (1.5,-1.5)
            {\includegraphics[width=3cm,height=3cm]{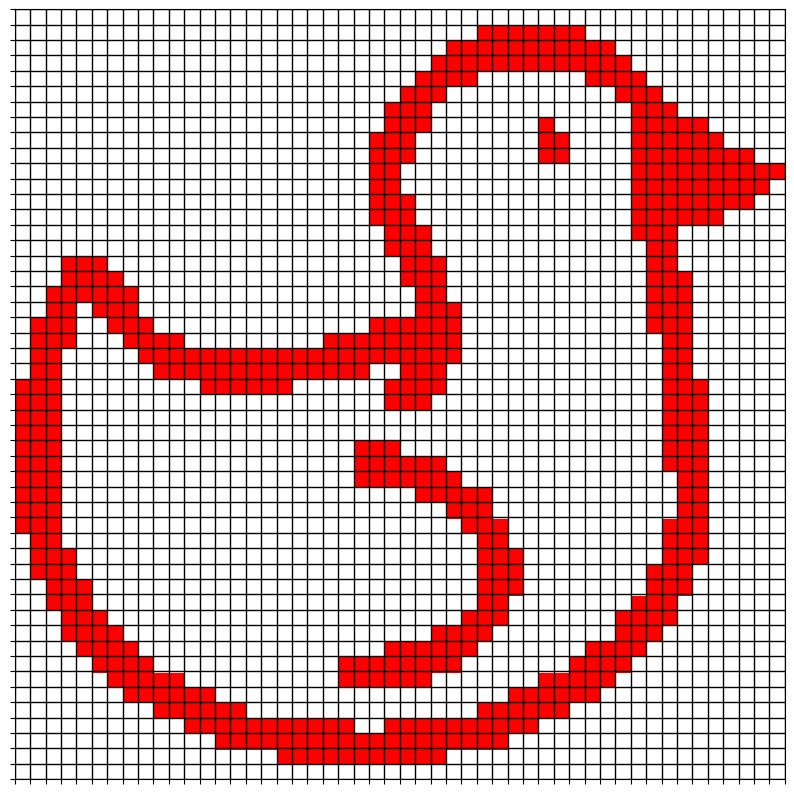}};
        \filldraw [green] (0.76,-1.32) rectangle (0.81, -1.37);
        \draw[black, thick, ->] (0.2,0.2) .. controls (0.5,-0.5) and (0.5,-0.5) .. (0.79,-1.33);
        \filldraw [black] (0.05,-2.95) circle (0.05);
        \draw[black, thick, ->] (0.05,-2.95) -- node[above, rotate=90] {$w_O + \beta \cdot w_\text{up}$} (0.05,0.0);
        \draw[black, thick, ->] (0.05,-2.95) -- node[below] {$w_O + \alpha \cdot w_\text{right}$} (3.0,-2.95);
        \node at (-0.2,-3.2) {$w_O$};
        \end{tikzpicture}
         \caption{Parametrization of the pattern downsampled to $50\times50$ size}
         \label{fig:method:duck-grid}
     \end{subfigure}
     \hfill
     \begin{subfigure}[b]{0.32\textwidth}
         \centering
         \begin{tikzpicture}
        \tikzset{>=latex}
        \tikzstyle{every node}=[font=\small]
        \node[font=\tiny] at (1.5,0.4) {$w_{28,13} = w_O + 28 \cdot w_\text{up} + 13 \cdot w_\text{right}$};
        \node[inner sep=0pt] (whitehead) at (1.83,-1.53)
            {\includegraphics[width=3.65cm,height=2.9cm]{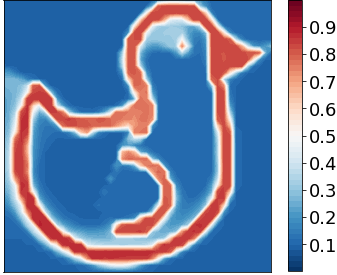}};
        \filldraw [green] (0.76,-1.32) rectangle (0.81, -1.37);
        \draw[black, thick, ->] (0.2,0.2) .. controls (0.5,-0.5) and (0.5,-0.5) .. (0.79,-1.33);
        \filldraw [black] (0.05,-2.95) circle (0.05);
        \draw[black, thick, ->] (0.05,-2.95) -- node[above, rotate=90] {$w_O + \beta \cdot w_\text{up}$} (0.05,0.0);
        \draw[black, thick, ->] (0.05,-2.95) -- node[below] {$w_O + \alpha \cdot w_\text{right}$} (3.0,-2.95);
        \node at (-0.2,-3.2) {$w_O$};
        \end{tikzpicture}
         \caption{Accuracy landscape of the found minimum with our method}
         \label{fig:method:fitting-result}
     \end{subfigure}
        \caption{Multi-point optimization method for 2D pattern fitting on FashionMNIST dataset.}
        \label{fig:method}
\end{figure}
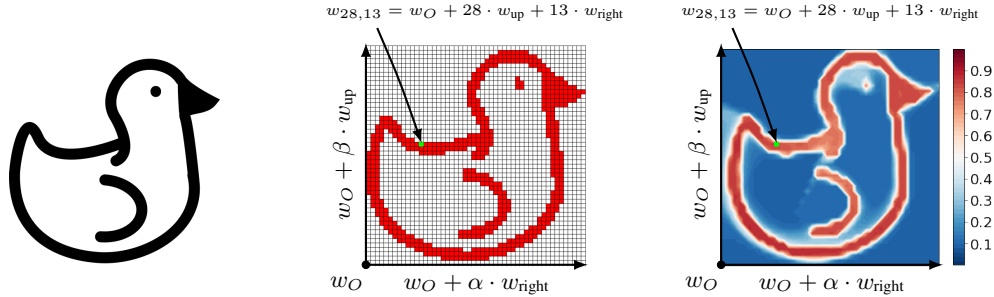

Now, imagine we want to find weight vectors $\bm w_1, ..., \bm w_K \in M_{\bm \theta}$ such that they form a pattern in the loss landscape depicted on figure \ref{fig:method:duck-grid}.
This means that they must lie on the same 2D-plane and be arranged in such a way that their contour map resembles the desired pattern, i.e. some weight vectors correspond to models with low loss values (\textit{black} pixels of the pattern) and others correspond to models with high loss values (\textit{white} pixels).
If one looks at a contour map computed around the arbitrary minimum, one generally sees a pit surrounded by random hills and mountains (\cite{Visualizing_Loss_Landscape}), but MPO allows to find a region with a nice-looking landscape, like on Figure \ref{fig:method:fitting-result}.

Consider our pattern contains $K$ pixels: this means that we want to find parameters of $K$ weight vectors $\bm w_1, ..., \bm w_K \in \mathbb{R}^n$ with pattern constraints discussed above.
Doing so without MPO would be a tedious process since $K$ can be very large (in our experiments we usually have $K = 50^2$).
To perform MPO we first parametrize our 2D-plane $M_{\bm \theta}$ with three weight vectors: $(\bm{w}_O, \bm{w}_\text{up}, \bm{w}_\text{right}) = \bm\theta \in \mathbb{R}^{3n}$.
Next, for each $\bm w_i$ we associate a pair of numbers $(\alpha, \beta) \in \mathbb{N}^2$, which correspond to $(\alpha, \beta)$-th pixel of the picture \ref{fig:method:duck-icon}, and which specifies its position on $M_{\bm \theta}$:
\begin{equation}
\label{eq:coordinates}
    \bm w_{i} = \bm{w}_{\alpha,\beta} = \bm{w}_O + \alpha \cdot \bm{w}_\text{right} + \beta \cdot \bm{w}_\text{up}.
\end{equation}
Since parametrization \eqref{eq:coordinates} is just a linear combination of some vectors we can compute the gradients of any differentiable loss function $\mathcal{L}(\bm{w}_{\alpha, \beta})$ with respect to $\bm \theta$.
And since for each model we have to keep only its 2D-coordinates instead of the corresponding large $n$-dimensional weight vector we obtain substantial memory savings.

We would like the coordinate system, defined by $\bm{w}_O, \bm{w}_\text{up}, \bm{w}_\text{right}$, to be orthogonal and properly scaled. 
For this we enforce $\bm{w}_\text{up} \perp \bm{w}_\text{right}$ and $\| \bm{w}_\text{up} \| = \| \bm{w}_\text{right} \|$.
We introduce a new parameter vector $\bm{\phi}_\text{right} \in \mathbb{R}^n$ instead of $\bm{w}_\text{right}$ and optimize for it.
Vector $\bm{w}_\text{right}$ is directly computed from $\bm{\phi}_\text{right}$ and $\bm{w}_\text{up}$ via Gram-Shmidt orthogonalization process:
\begin{equation}
    \hat{\bm w}_\text{right} = \bm{\phi}_\text{right} - \frac{\langle \bm{\phi}_\text{right}, \bm{w}_\text{up} \rangle}{\| \bm{w}_\text{up} \|^2} \cdot \bm{w}_\text{up},
    \qquad
    \bm{w}_\text{right} = \frac{\| \hat{\bm w}_\text{up} \|}{\| \hat{\bm w}_\text{right} \|} \hat{\bm w}_\text{right}.
\end{equation}
One can easily note that it is differentiable with respect to $\bm{\phi}_\text{right}$, that's why we have no trouble to optimize it via gradient methods.
To fit a pattern we minimize the cross-entropy in black points and maximize it in white points, i.e. we optimize the following functional:
\begin{equation}\label{eq:functional}
    \mathcal{L}(\bm \theta) = \frac{1}{T}\sum_{t=1}^T\left[\frac{1}{|P_+|}\sum_{(\alpha, \beta) \in P_+} \log p_{\bm w_{\alpha, \beta}}(\bm y^{(t)} | \bm x^{(t)}) - \frac{1}{|P_-|} \sum_{(\alpha, \beta) \in P_-} \log p_{\bm w_{\alpha, \beta}}(\bm y^{(t)} | \bm x^{(t)})\right],
\end{equation}
where $P_-, P_+$ are sets of indices of black and white pixels of a pattern respectively, $\{(\bm x^{(t)}, \bm y^{(t)})\}_{t=1}^T$ are training pairs and $p_{\bm w_{\alpha, \beta}}(\bm y | \bm x)$ is our model at a point $\bm w_{\alpha, \beta}$.
We optimize for $\bm\theta$ with standard gradient methods and the details of this procedure are specified in appendix \ref{appendix:optimization}.


\section{Experiments}
\subsection{Finding patterns of the loss landscape}

Intuitively, a loss surface with irregular and complicated landscape should be hard to optimize.
In the first series of experiments, we explore how diverse and sophisticated it can be by searching for different patterns in the loss landscape of neural networks.
We take a simple VGG-like CNN architecture (described in appendix \ref{appendix:hyperparams}) and optimize the parameters of a 2D-plane described in section \ref{section:method} on FashionMNIST and CIFAR10 datasets to find a specific pattern.

To our surprise, literally, any pattern of an adequate size that we tried was found with high quality in the loss landscape.
Heat maps of the found 2D-planes are presented in Figure \ref{fig:icons-grid} and in appendix \ref{appendix:visualizations}.
These results demonstrate that the loss surface of neural networks can be very curved and ill-behaved.
Interestingly, for FashionMNIST dataset resulting patterns evaluated on a train set are almost indistinguishable from the patterns evaluated on a test set.
This indicates that loss surface diversity is a property of the whole risk functional, not only its empirical counterpart.

\subsection{Smoothing properties of batch normalization}

\begin{figure}
    \centering
    \begin{subfigure}{0.3\textwidth}
        \centering
        \includegraphics[height=2.5cm]{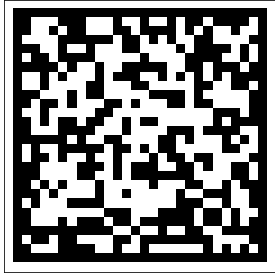}
        \caption{Mask of the pattern}
        \label{fig:random-squares:mask}
    \end{subfigure}
    \begin{subfigure}{0.3\textwidth}
        \centering
        \includegraphics[height=2.5cm]{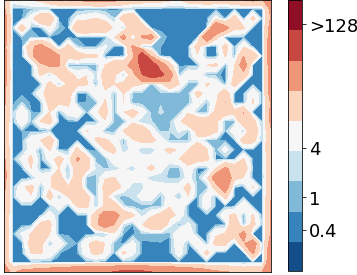}
        \caption{Without batch normalization}
        \label{fig:random-squares:without-bn}
    \end{subfigure}
    \begin{subfigure}{0.3\textwidth}
        \centering
        \includegraphics[height=2.5cm]{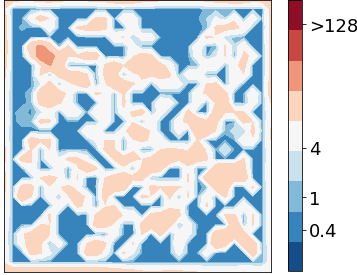}
        \caption{With batch normalization}
        \label{fig:random-squares:with-bn}
    \end{subfigure}
    \caption{(a) Example of a random binary mask with a filling probability of 0.5. (b) The result of the MPO procedure for the model without batch normalization. (c) The result of the MPO procedure for the model with batch normalization.}
\end{figure}

In the second batch of experiments, we study the smoothing properties of batch normalization.
For this, we generate random binary masks of size $30 \times 30$.
We vary the probability of pixels being masked from 0.1 to 0.9, thus obtaining the patterns of varying complexity.
For the probability of 0.1 and 0.9, the landscape is mostly homogeneous, being either completely black or completely white.
For probabilities near 0.5, we get very irregular patterns that are very difficult to fit (see  Figure \ref{fig:random-squares:mask}).

For each random pattern, we've trained two types of CNN models on FashionMNIST dataset, one with batch normalization and another without it.
We then measured the mean accuracy in black and white pixels on a test set and took the difference.
The difference in accuracy across 5 runs with the corresponding standard deviation is depicted in Figure \ref{fig:batch-norm-acc-diff}.
As one can see, it's much more difficult for a model with batch normalization to fit the desired pattern.
This indicates that such complex patterns as random binary masks are less likely to occur on its loss surface, implying that it is more smooth and regular.
And as Figure \ref{fig:batch-norm-mean-accs} shows models trained with batch normalization generally attain higher scores --- a consequence of a more decent loss landscape.

We note that batch normalization is not well aligned with our parametrization of 2D-plane, so we used a small trick to alleviate this.
The problem and the solution are discussed in appendix \ref{appendix:bn}.

\begin{figure}
    \centering
    \begin{subfigure}[t]{0.49\textwidth}
        \centering
\begin{tikzpicture}

\begin{axis}[
width=7cm,
height=5cm,
axis line style={white},
legend cell align={left},
legend style={draw=white!80.0!black, fill=white!89.80392156862746!black},
tick align=outside,
tick pos=left,
xlabel={Filling probability},
xmajorgrids,
xmin=0.06, xmax=0.94,
xtick style={color=white!33.33333333333333!black},
ylabel={Accuracy difference},
ymajorgrids,
ymin=0.17835696312977, ymax=0.51349512786398,
ytick style={color=white!33.33333333333333!black}
]
\path [fill=blue, fill opacity=0.1]
(axis cs:0.1,0.425684439139905)
--(axis cs:0.1,0.498261574921516)
--(axis cs:0.2,0.440394576989444)
--(axis cs:0.3,0.391437991487654)
--(axis cs:0.4,0.364494976485688)
--(axis cs:0.5,0.330930173923867)
--(axis cs:0.6,0.321447431583623)
--(axis cs:0.7,0.332045929859005)
--(axis cs:0.8,0.334532071547089)
--(axis cs:0.9,0.376523356360745)
--(axis cs:0.9,0.308982474105506)
--(axis cs:0.9,0.308982474105506)
--(axis cs:0.8,0.287616265726467)
--(axis cs:0.7,0.302451601364732)
--(axis cs:0.6,0.287730223651604)
--(axis cs:0.5,0.288866687184848)
--(axis cs:0.4,0.3161198239085)
--(axis cs:0.3,0.345420674430864)
--(axis cs:0.2,0.388500710503587)
--(axis cs:0.1,0.425684439139905)
--cycle;

\path [fill=red, fill opacity=0.1]
(axis cs:0.1,0.294480463067547)
--(axis cs:0.1,0.321045653443264)
--(axis cs:0.2,0.284422020374301)
--(axis cs:0.3,0.266231699836555)
--(axis cs:0.4,0.244605639188131)
--(axis cs:0.5,0.247912073472134)
--(axis cs:0.6,0.236585830880898)
--(axis cs:0.7,0.241268020555081)
--(axis cs:0.8,0.247700802633891)
--(axis cs:0.9,0.277834501609621)
--(axis cs:0.9,0.20973222642619)
--(axis cs:0.9,0.20973222642619)
--(axis cs:0.8,0.197726254632717)
--(axis cs:0.7,0.195069485024874)
--(axis cs:0.6,0.205075145401999)
--(axis cs:0.5,0.193590516072234)
--(axis cs:0.4,0.202210924332594)
--(axis cs:0.3,0.234543217788878)
--(axis cs:0.2,0.263071408849859)
--(axis cs:0.1,0.294480463067547)
--cycle;

\addplot [semithick, blue]
table {%
0.1 0.46197300703071
0.2 0.414447643746516
0.3 0.368429332959259
0.4 0.340307400197094
0.5 0.309898430554357
0.6 0.304588827617613
0.7 0.317248765611868
0.8 0.311074168636778
0.9 0.342752915233125
};
\addlegendentry{No batch normalization}
\addplot [semithick, red]
table {%
0.1 0.307763058255406
0.2 0.27374671461208
0.3 0.250387458812716
0.4 0.223408281760363
0.5 0.220751294772184
0.6 0.220830488141449
0.7 0.218168752789978
0.8 0.222713528633304
0.9 0.243783364017905
};
\addlegendentry{With batch normalization}
\end{axis}

\end{tikzpicture}
        \caption{The difference in accuracy. Standard deviations are taken across 5 runs.}
        \label{fig:batch-norm-acc-diff}
    \end{subfigure}
    \hfill
    \begin{subfigure}[t]{0.49\textwidth}
        \centering
\begin{tikzpicture}

\begin{axis}[
width=7cm,
height=5cm,
axis line style={white},
legend cell align={left},
legend style={at={(0.97,0.03)}, anchor=south east, draw=white!80.0!black, fill=white!89.80392156862746!black},
tick align=outside,
tick pos=left,
xlabel={Filling probability},
xmajorgrids,
xmin=0.06, xmax=0.94,
xtick style={color=white!33.33333333333333!black},
ylabel={Mean accuracy},
ymajorgrids,
ymin=0.268670363466878, ymax=0.887699632752529,
ytick style={color=white!33.33333333333333!black}
]
\path [fill=blue, fill opacity=0.1]
(axis cs:0.1,0.777890572521251)
--(axis cs:0.1,0.812589808933863)
--(axis cs:0.2,0.772734134040635)
--(axis cs:0.3,0.753727150722824)
--(axis cs:0.4,0.738541075746978)
--(axis cs:0.5,0.722617714284207)
--(axis cs:0.6,0.711602298751627)
--(axis cs:0.7,0.747298727615049)
--(axis cs:0.8,0.761326078036125)
--(axis cs:0.9,0.824896609633584)
--(axis cs:0.9,0.788445761766329)
--(axis cs:0.9,0.788445761766329)
--(axis cs:0.8,0.733789721562227)
--(axis cs:0.7,0.728561481717537)
--(axis cs:0.6,0.705046222099195)
--(axis cs:0.5,0.704471870227522)
--(axis cs:0.4,0.702910025769416)
--(axis cs:0.3,0.720356428929019)
--(axis cs:0.2,0.750406440657159)
--(axis cs:0.1,0.777890572521251)
--cycle;

\path [fill=blue, fill opacity=0.1]
(axis cs:0.1,0.296808057525317)
--(axis cs:0.1,0.369726309868377)
--(axis cs:0.2,0.363209639772941)
--(axis cs:0.3,0.392347743281297)
--(axis cs:0.4,0.398551259518024)
--(axis cs:0.5,0.429799711853131)
--(axis cs:0.6,0.420416257114505)
--(axis cs:0.7,0.443903483471085)
--(axis cs:0.8,0.470320690818128)
--(axis cs:0.9,0.49645258479294)
--(axis cs:0.9,0.431383956140723)
--(axis cs:0.9,0.431383956140723)
--(axis cs:0.8,0.402646771506668)
--(axis cs:0.7,0.397459194637764)
--(axis cs:0.6,0.38705460850109)
--(axis cs:0.5,0.377493011549883)
--(axis cs:0.4,0.362285041604183)
--(axis cs:0.3,0.344877170452027)
--(axis cs:0.2,0.331035647431821)
--(axis cs:0.1,0.296808057525317)
--cycle;

\path [fill=red, fill opacity=0.1]
(axis cs:0.1,0.819131271121412)
--(axis cs:0.1,0.837587613197806)
--(axis cs:0.2,0.815782034824135)
--(axis cs:0.3,0.804523404728565)
--(axis cs:0.4,0.80141535225123)
--(axis cs:0.5,0.79994966803316)
--(axis cs:0.6,0.810275586066576)
--(axis cs:0.7,0.818204247176996)
--(axis cs:0.8,0.836414571626695)
--(axis cs:0.9,0.85956193869409)
--(axis cs:0.9,0.84749528297761)
--(axis cs:0.9,0.84749528297761)
--(axis cs:0.8,0.821118520815743)
--(axis cs:0.7,0.806842751864106)
--(axis cs:0.6,0.791012820089637)
--(axis cs:0.5,0.785079731373566)
--(axis cs:0.4,0.794031911021303)
--(axis cs:0.3,0.794651287646504)
--(axis cs:0.2,0.80133654623186)
--(axis cs:0.1,0.819131271121412)
--cycle;

\path [fill=red, fill opacity=0.1]
(axis cs:0.1,0.507609746829455)
--(axis cs:0.1,0.533583020978952)
--(axis cs:0.2,0.548081877294667)
--(axis cs:0.3,0.566214766395101)
--(axis cs:0.4,0.594950162817727)
--(axis cs:0.5,0.598538554626314)
--(axis cs:0.6,0.591668937648717)
--(axis cs:0.7,0.619980701012214)
--(axis cs:0.8,0.637505594426417)
--(axis cs:0.9,0.648970566002349)
--(axis cs:0.9,0.57051992763354)
--(axis cs:0.9,0.57051992763354)
--(axis cs:0.8,0.574600440749413)
--(axis cs:0.7,0.568728792448933)
--(axis cs:0.6,0.567958492224598)
--(axis cs:0.5,0.544988255236045)
--(axis cs:0.4,0.553680536934081)
--(axis cs:0.3,0.532185008354536)
--(axis cs:0.2,0.521543274537168)
--(axis cs:0.1,0.507609746829455)
--cycle;

\addplot [semithick, blue]
table {%
0.1 0.795240190727557
0.2 0.761570287348897
0.3 0.737041789825921
0.4 0.720725550758197
0.5 0.713544792255864
0.6 0.708324260425411
0.7 0.737930104666293
0.8 0.747557899799176
0.9 0.806671185699956
};
\addlegendentry{No BN, black}
\addplot [semithick, blue, dashed]
table {%
0.1 0.333267183696847
0.2 0.347122643602381
0.3 0.368612456866662
0.4 0.380418150561103
0.5 0.403646361701507
0.6 0.403735432807797
0.7 0.420681339054425
0.8 0.436483731162398
0.9 0.463918270466831
};
\addlegendentry{No BN, white}
\addplot [semithick, red]
table {%
0.1 0.828359442159609
0.2 0.808559290527997
0.3 0.799587346187535
0.4 0.797723631636267
0.5 0.792514699703363
0.6 0.800644203078107
0.7 0.812523499520551
0.8 0.828766546221219
0.9 0.85352861083585
};
\addlegendentry{With BN, black}
\addplot [semithick, red, dashed]
table {%
0.1 0.520596383904204
0.2 0.534812575915918
0.3 0.549199887374819
0.4 0.574315349875904
0.5 0.571763404931179
0.6 0.579813714936658
0.7 0.594354746730574
0.8 0.606053017587915
0.9 0.609745246817945
};
\addlegendentry{With BN, white}
\end{axis}

\end{tikzpicture}
        \caption{Mean accuracy for different values of filling probability. Blue --- without and red --- with batch normalization. A solid line --- for black pixels and a dotted one --- for white pixels.}
        \label{fig:batch-norm-mean-accs}
    \end{subfigure}
\end{figure}

\section{Conclusion}

We presented a method that allows one to fit several points in the parameter space by running an optimization procedure on a significantly smaller amount of parameters. 
It was used to demonstrate that the loss surface of neural networks conceals regions with arbitrary landscape patterns, implicating its diversity and optimization complexity.
Besides, along the way, we showed that batch normalization makes it more regular.
It is important to note, however, that though our approach provides drastic memory savings, it does not give notable benefits in terms of clock-wall time since on each iteration we perform many forward passes to update the parameters of the underlying manifold.
In current submission, we didn't explore its use to construct better, more decorrelated ensembles, but we believe it to be a very fruitful research direction with potential practical applications.
Theoretical analysis was also left for future work.

\subsubsection*{Acknowledgments}

The work was supported by National Technology Initiative and PAO Sberbank project ID 0000000007417F630002.
All the icons used in our work were taken from www.flaticon.com.
We thank Eugene Golikov and Mikhail Arkhipov for valuable discussions and suggestions.

\small

\bibliography{references}

\appendix

\clearpage
\section{Optimization details}
\label{appendix:optimization}
First, we downsample an image to a size $w \times h$ (usually $50 \times 50$). Then we initialize the parameters of each layer for weights $\bm{w}_O, \bm{w}_\text{up}$ and $\phi_\text{right}$ via Xavier initialization.
We also found it beneficial to scale vectors $\bm{w}_\text{up}, \bm{w}_\text{right}$ by some small scale value $s$, initializing it somewhere between 0.01 and 0.1 depending on the size of a pattern.
We found it important to optimize the scaling factor $s \in \mathbb{R}$ as well, since adjusting it manually is a difficult process which requires extensive hyperparameters tuning.
So finally we minimize \eqref{eq:functional} for $\bm\theta = \{\bm{w}_O, \bm{w}_\text{up}, \bm{\phi}_\text{right}, s \}$ via Adam optimizer.
On each iteration, we pick random subsets of $P_-$ and $P_+$ to compute the gradients to make the training procedure faster.

Also during training, we do not use the pixels which are surrounded by the pixels of the same class to reduce computation since we found that model interpolates between neighboring values without being trained to do so.

Batch Normalization will not work for this procedure as it is, because during orthogonalization process scale parameters become negative.
To make it valid we add $-0.5$ for its scale parameter to make it have zero mean and add $0.5$ back during the forward pass.
Details about this trick are provided in the appendix \ref{appendix:bn}.



\section{Experiment details}
\label{appendix:hyperparams}
For experiments for finding pictures in the loss landscape, our architecture consisted of 3 convolutional layers with kernel sizes of 3 and 8, 32, 64 number of channels and ReLU non-linearity after each layer.
Then we used adaptive average pooling to convert the representation to $64\times4\times4$ tensor, flattened it and passed to a one-layer MLP with 128 hidden units.

For experiments with batch normalization we used a very similar architecture, the only difference was that we had more convolutional layers with 8, 8, 32, 32, 64, 64 number of channels.
We had batch normalization after each convolutional layer (i.e. before activation is applied) for a model with batch normalization.

We used Adam optimizer with a learning rate of 0.0003.
We thresholded cross-entropy value in white cells by a value of 2.5: otherwise it was too easy for the model just to have a very large cross-entropy value everywhere to get a good optimization loss value.
We trained the model with a batch size of 512 and used 50 cells per update in each iteration.
We initialize scale value $s$ at 0.1.

\section{Batch Normalization reparametrization}
\label{appendix:bn}
Consider we have two random vectors $\bm u, \bm v \in \mathbb{R}^n$ and we want to perform a Gram-Shmidt orthogonalization to get vector $\bm w$ such that $\bm w \perp \bm u$.
If $\bm u, \bm v$ come from zero-mean normal distribution with a small variance, their elements are i.i.d and $n$ is large enough, then $\langle \bm u, \bm v \rangle$ will have zero mean with a small variance.
Such an initialization scheme is quite typical for most of the layers, but not for normalization ones, like Batch Normalization.
This is because scale parameters $\bm s$ of BN layer are initialized from $U[0,1]$ (or even all ones).
As a result, the scalar product of the parameters of two BN layers is much larger than zero, which, in turn, results in vector $\bm w$ having a lot of negative values:
\begin{equation}
    \bm w = \bm v - \frac{\langle \bm v, \bm u \rangle}{\| \bm u \|^2} \bm u
\end{equation}
Such a behavior is not fatal, but not desirable, because scale parameter associates with standard deviation, so this is an abnormal situation when it's negative.

To alleviate this issue we parametrize batch normalization in such a way that scale parameter has zero mean.
We do this simply by initializing it from $U[-0.5, 0.5]$ instead of $U[0, 1]$.
Since this makes it contain negative values, during the forward pass we have to add 0.5 back to make it all-positive again.

\clearpage
\section{Additional visualizations of loss and accuracy surfaces}
\label{appendix:visualizations}

\begin{figure}[H]
\centering
\begin{subfigure}[b]{0.49\textwidth}
    \centering
    \includegraphics[height=2.3cm]{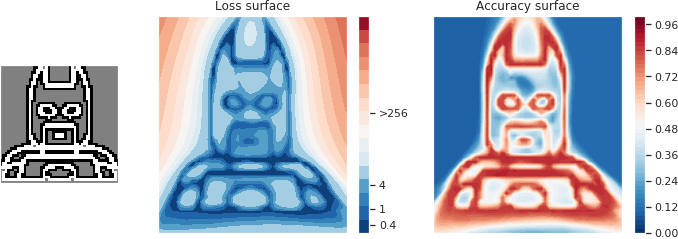}
\end{subfigure}
\hfill
\begin{subfigure}[b]{0.49\textwidth}
    \centering
    \includegraphics[height=2.3cm]{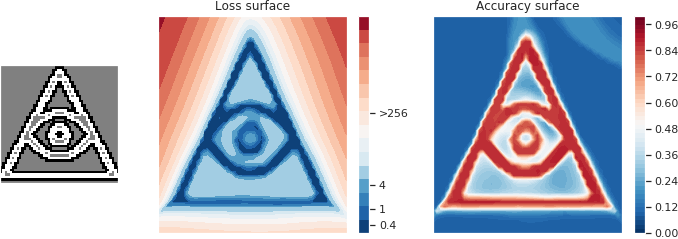}
\end{subfigure}

\begin{subfigure}[b]{0.49\textwidth}
    \centering
    \includegraphics[height=2.3cm]{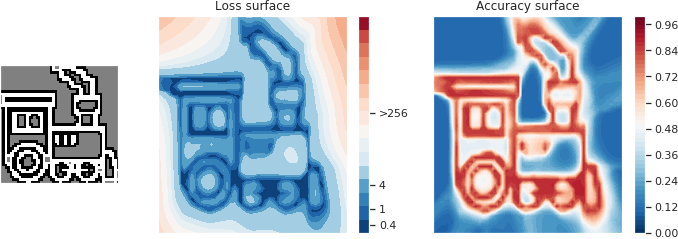}
\end{subfigure}
\hfill
\begin{subfigure}[b]{0.49\textwidth}
    \centering
    \includegraphics[height=2.3cm]{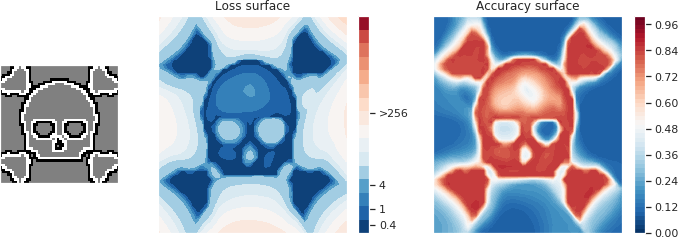}
\end{subfigure}

\begin{subfigure}[b]{0.49\textwidth}
    \centering
    \includegraphics[height=2.3cm]{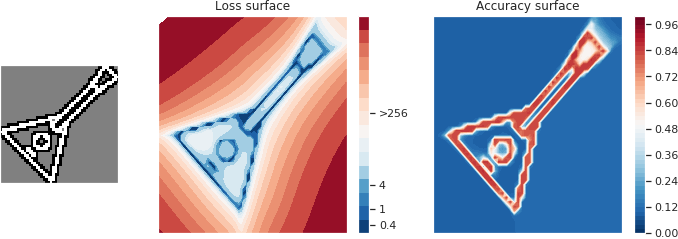}
\end{subfigure}
\hfill
\begin{subfigure}[b]{0.49\textwidth}
    \centering
    \includegraphics[height=2.3cm]{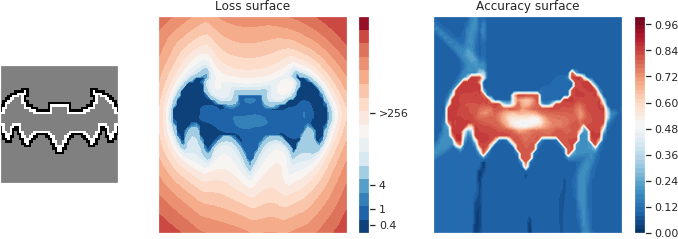}
\end{subfigure}

\begin{subfigure}[b]{0.49\textwidth}
    \centering
    \includegraphics[height=2.3cm]{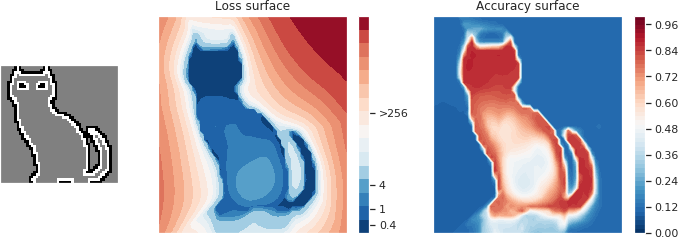}
\end{subfigure}
\hfill
\begin{subfigure}[b]{0.49\textwidth}
    \centering
    \includegraphics[height=2.3cm]{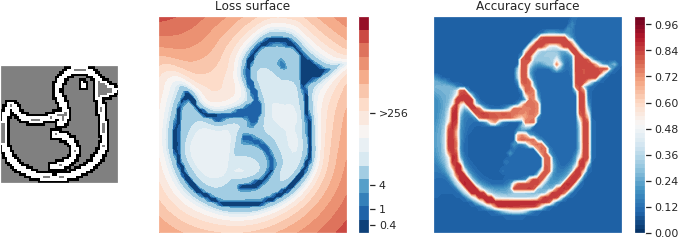}
\end{subfigure}

\begin{subfigure}[b]{0.49\textwidth}
    \centering
    \includegraphics[height=2.3cm]{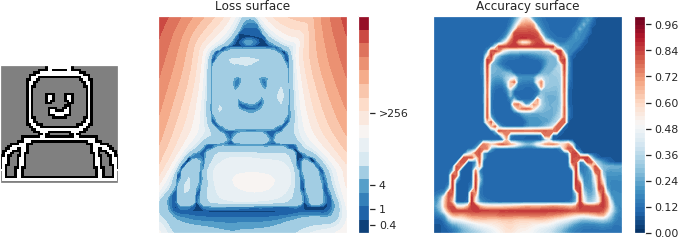}
\end{subfigure}
\hfill
\begin{subfigure}[b]{0.49\textwidth}
    \centering
    \includegraphics[height=2.3cm]{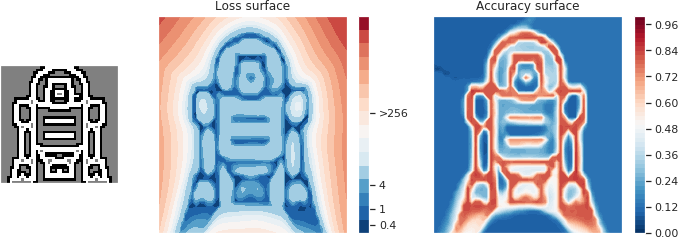}
\end{subfigure}

\begin{subfigure}[b]{0.49\textwidth}
    \centering
    \includegraphics[height=2.3cm]{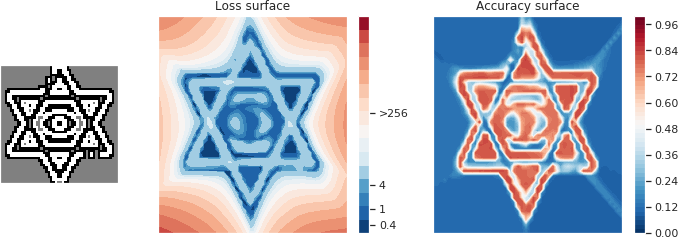}
\end{subfigure}
\hfill
\begin{subfigure}[b]{0.49\textwidth}
    \centering
    \includegraphics[height=2.3cm]{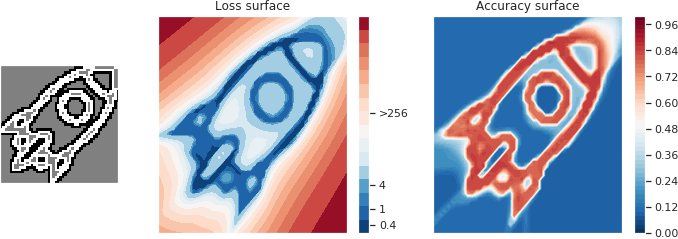}
\end{subfigure}

\begin{subfigure}[b]{0.49\textwidth}
    \centering
    \includegraphics[height=2.3cm]{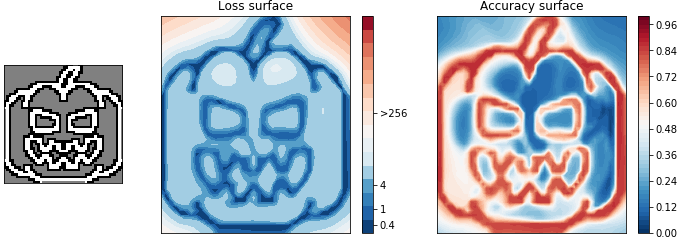}
\end{subfigure}
\hfill
\begin{subfigure}[b]{0.49\textwidth}
    \centering
    \includegraphics[height=2.3cm]{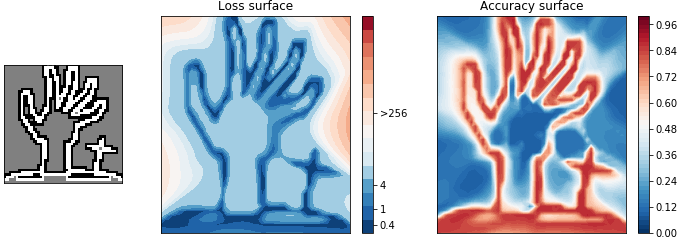}
\end{subfigure}
\caption{Additional results for pattern search on FashionMNIST dataset. Since train and test landscapes are almost visually indistinguishable for our model in the case of FashionMNIST dataset, we depict here only \textit{test} loss surfaces.}
\end{figure}

\clearpage

\begin{figure}[H]
\centering
\begin{subfigure}[b]{0.49\textwidth}
    \centering
    \includegraphics[height=2.3cm]{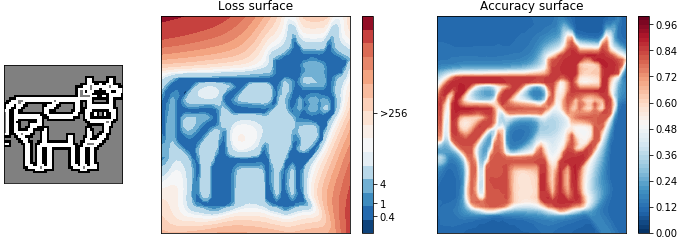}
\end{subfigure}
\hfill
\begin{subfigure}[b]{0.49\textwidth}
    \centering
    \includegraphics[height=2.3cm]{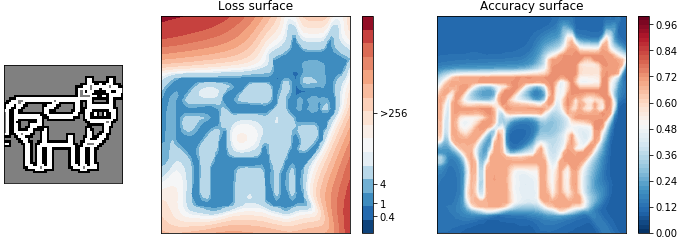}
\end{subfigure}

\begin{subfigure}[b]{0.49\textwidth}
    \centering
    \includegraphics[height=2.3cm]{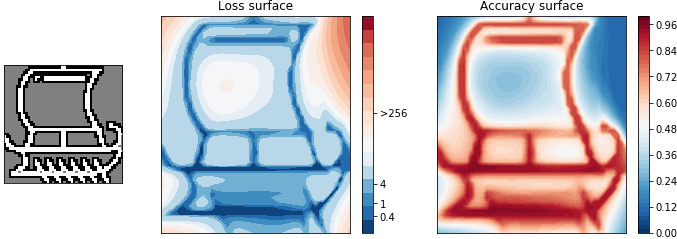}
\end{subfigure}
\hfill
\begin{subfigure}[b]{0.49\textwidth}
    \centering
    \includegraphics[height=2.3cm]{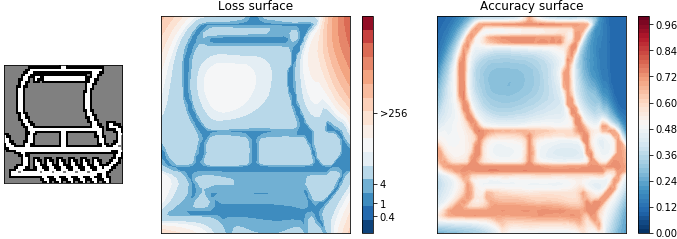}
\end{subfigure}

\begin{subfigure}[b]{0.49\textwidth}
    \centering
    \includegraphics[height=2.3cm]{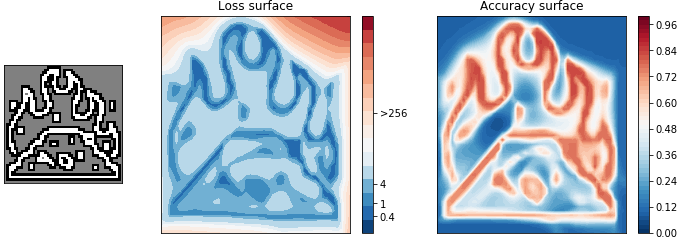}
\end{subfigure}
\hfill
\begin{subfigure}[b]{0.49\textwidth}
    \centering
    \includegraphics[height=2.3cm]{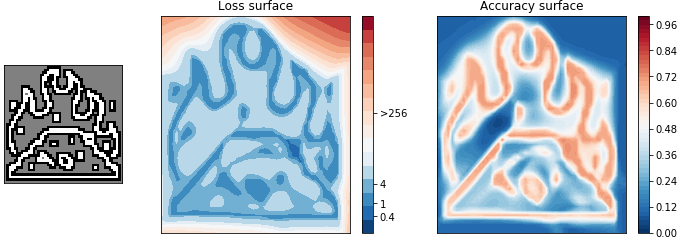}
\end{subfigure}

\begin{subfigure}[b]{0.49\textwidth}
    \centering
    \includegraphics[height=2.3cm]{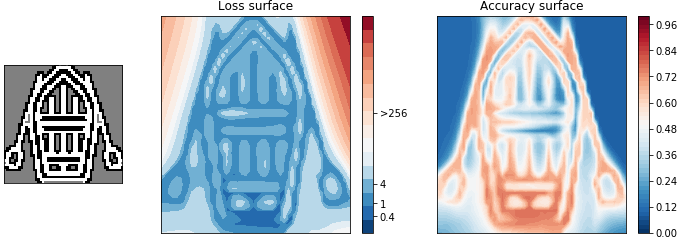}
\end{subfigure}
\hfill
\begin{subfigure}[b]{0.49\textwidth}
    \centering
    \includegraphics[height=2.3cm]{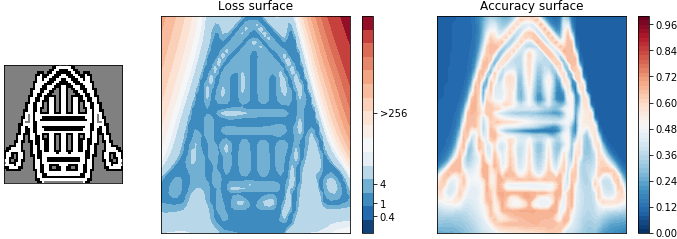}
\end{subfigure}

\begin{subfigure}[b]{0.49\textwidth}
    \centering
    \includegraphics[height=2.3cm]{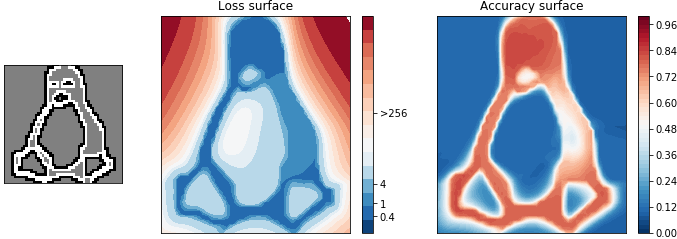}
\end{subfigure}
\hfill
\begin{subfigure}[b]{0.49\textwidth}
    \centering
    \includegraphics[height=2.3cm]{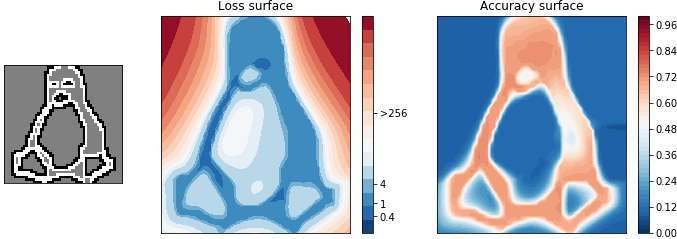}
\end{subfigure}

\begin{subfigure}[b]{0.49\textwidth}
    \centering
    \includegraphics[height=2.3cm]{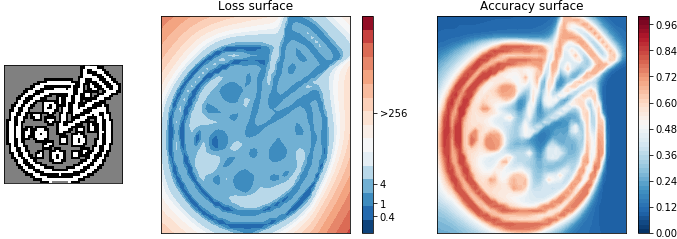}
\end{subfigure}
\hfill
\begin{subfigure}[b]{0.49\textwidth}
    \centering
    \includegraphics[height=2.3cm]{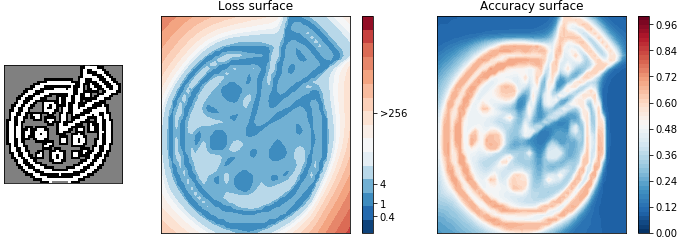}
\end{subfigure}
\caption{Additional results for pattern search on CIFAR10 dataset. Left column depicts train loss/accuracy surface, right column depicts test loss/accuracy surface.}
\end{figure}

\end{document}